\documentclass[12pt]{article}

\usepackage{sbc-template}
\usepackage{amssymb}
\usepackage{graphicx,xurl}

\usepackage[utf8]{inputenc}  
\usepackage{adjustbox}
\usepackage{xcolor}

\usepackage{amsmath}

\graphicspath{{pics/}}

\usepackage{array}
\newcolumntype{P}[1]{>{\centering\arraybackslash}p{#1}}
     
\title{Face Reconstruction with Variational Autoencoder and Face Masks}

\author{Rafael S. Toledo\inst{1} \and Eric A. Antonelo\inst{1}}

\address{Department of Automation and Systems \\
         Federal University of Santa Catarina (UFSC), Florianópolis, Brazil
    \email{rafael.toledo@posgrad.ufsc.br, eric.antonelo@ufsc.br}
}

\begin{document} 

\maketitle

\begin{abstract}
Variational AutoEncoders (VAE) employ deep learning models to learn a continuous latent z-space that is subjacent to a high-dimensional observed dataset. With that, many tasks are made possible, including face reconstruction and face synthesis.
In this work, we investigated how face masks can help the training of VAEs for face reconstruction, by restricting the learning to the pixels selected by the face mask. 
An evaluation of the proposal using the celebA dataset shows that the reconstructed images are enhanced with the face masks, especially when SSIM loss is used either with $l_1$ or $l_2$ loss functions.
We noticed that the inclusion of a decoder for face mask prediction in the architecture affected the performance for $l_1$ or $l_2$ loss functions, while this was not the case for the SSIM loss.
Besides, SSIM perceptual loss yielded the crispest samples between all hypotheses tested, although it shifts the original color of the image, making the usage of the $l_1$ or $l_2$ losses together with SSIM helpful to solve this issue.\footnote{Code and models are available on \url{https://github.com/tldrafael/FaceReconstructionWithVAEAndFaceMasks}.}
\end{abstract}

\section{Introduction}
%

Deep Generative Models became extremely popular recently for image synthesis, generation, and manipulation \cite{snell2017learning,qian2019make,dosovitskiy2016generating,larsen2016autoencoding,esser2018variational}. One of the major challenges in this area corresponds to designing models capable of yielding photo-realistic faces, which would be useful for many industries like films, games, photograph editions, or even face anonymization.

The most popular deep learning approaches to tackle face synthesis and manipulation derive from Generative Adversarial Nets (GAN) \cite{goodfellow2014generative}, and Variational Autoencoders (VAE) \cite{kingma2014auto, rezende2014stochastic}. Currently, the state-of-the-art performance and ability to render crisper samples are obtained by GANs \cite{brock2018large, liu2019stgan, he2019attgan}. It is common sense that VAE usually presents blurriness on the rendered images \cite{dai2019diagnosing}. 

Nonetheless, there are many good reasons to keep improving the VAE framework in the image reconstruction and synthesis realm. For example, while GANs can generate images of high subjective perceptual quality, they are not able to fully capture the diversity of the true distribution \cite{razavi2019generating}, lacking diversity when synthesizing new samples; this is a problem known as mode collapse \cite{mescheder2018training}. Besides, GANs tend to lack full support over the data as opposed to the likelihood-based generative models from VAEs, which are better density models in terms of the likelihood criterion \cite{kingma2019introduction}.

Other drawbacks about GANs are that they are hard to optimize due to the lack of a closed-form loss function, and they can generate visually absurd outputs \cite{khan2018adversarial}.  On the other hand, VAEs possess some desirable properties like stable training, an interpretable encoder/inference network, outlier-robustness \cite{dai2019diagnosing}, and its paradigm allows an exploration of the disentangling intrinsic facial properties in the latent space \cite{qian2019make}.

Along with the blurry issue related to VAE, another cause of blurriness comes from training with $l_2$ or $l_1$ losses functions. The $l_2$ loss function is known to not characterize well the perceived visual quality of images  \cite{zhang2012comprehensive, wang2009mean,zhao2016loss}. 
This is because the $l_2$-norm of the error assumes pixel-wise independence, which is not true for natural images. For example, blurring causes a large human visual perceptual error but a low $l_2$ error change \cite{zhang2018unreasonable}.

To help to correct it, we tested the usage of the Structural Similarity Index (SSIM) \cite{wang2004image} as a loss function. SSIM is based on the structural information that compares local patterns of pixel intensities that have been normalized for luminance and contrast \cite{wang2009mean}. SSIM does not only compare pixels values independently between two images, but it also regards the pixel's neighbors values. SSIM compares the pixels in three aspects: luminance, contrast, and structure. 

Our proposal intended to attenuate these two factors of blurring, first adding a face mask based architecture that will help the learning process to focus only on the region of interest, and check other alternatives beyond the usual $l_n$ norms like SSIM and combinations of SSIM with $l_n$ norms. 

Our contributions are an architecture that helps the NN to ignore the background information to improve the performance on the face reconstruction task; an investigation on the usage of three popular metrics $l_1$, $l_2$, and SSIM; besides an investigation on how the combine these three losses functions to amend their individual drawbacks. In the end, it is shown that the proposed architecture plus a combined loss function worked better for the face reconstruction task.

\section{Background}
%
\subsection{Variational Autoencoder}
Deep Latent Variable Model (DLVM) are models whose latent variables distributions are parametrized by a neural network, e.g. $p_\theta(\mathbf{x},\mathbf{z})$. Latent variables commonly denoted as $z$ are variables which are part of the model but cannot be directly observed.  On a DLVM, we want to learn the true distribution of the data $p^*(x)$ by learning the marginal distribution of $p(\mathbf{x}) = \int_{z} p_\theta(\mathbf{x},\mathbf{z})\mathrm{d}z$. But, $p_\theta(\mathbf{x},\mathbf{z})$ or $p_\theta(\mathbf{z}|\mathbf{x})$ are computational intractable \cite{kingma2019introduction}. They shall be estimated by the VAE framework.

VAE provides a computationally efficient way for optimizing DLVM jointly with a corresponding inference model using Stochastic Gradient Descent (SGD) \cite{kingma2019introduction}. The intractable posterior is estimated by the encoder $q_{\phi}(\mathbf{z}|\mathbf{x})$.

The VAE maximizes the ELBO (Evidence Lower Bound, also called Variational Lower Bound) of the data log-likelihood. The most common way to derive ELBO is using Jensen's inequality, which states $f(\mathbb{E}[X]) \le \mathbb{E}[f(X)]$, for any convex function $f$ as the $\log$ function of our case.

We can decompose the log-likelihood of the data as:

\begin{align}
\label{eqn:vae1}
\log{p_\theta(\mathbf{x})} & = \log{\int_{\mathbf{z}} p_\theta(\mathbf{x},\mathbf{z}) \mathrm{d}\mathbf{z}} \\
\label{eqn:vae2}
& = \log{\left[ \mathbb{E}_{\mathbf{z}\sim q_\phi(\mathbf{z}|\mathbf{x})}\left[ \frac{p_\theta(\mathbf{x},\mathbf{z})}{q_\phi(\mathbf{z}|\mathbf{x})} \right] \right]} \\
\intertext{\indent Applying Jensen's inequality in the eq. \ref{eqn:vae2}:}
\label{eqn:vae3}
\log{p_\theta(\mathbf{x})} & \ge \mathbb{E}_{\mathbf{z}\sim q_\phi(\mathbf{z}|\mathbf{x})}\log{\left[ \frac{p_\theta(\mathbf{x},\mathbf{z})}{q_\phi(\mathbf{z}|\mathbf{x})} \right]} \\
\label{eqn:vae4}
& = \mathbb{E}_{\mathbf{z}\sim q_\phi(\mathbf{z}|\mathbf{x})}\left[ \log{p_\theta(\mathbf{x},\mathbf{z})} \right] - \mathbb{E}_{\mathbf{z}\sim q_\phi(\mathbf{z}|\mathbf{x})}\left[ \log{q_\phi(\mathbf{z}|\mathbf{x})} \right]
\intertext{\indent Decomposing the joint distribution $p_\theta(\mathbf{x},\mathbf{z})$ to $p(\mathbf{z})p_\theta(\mathbf{x}|\mathbf{z})$:}
\label{eqn:vae5}
\log{p_\theta(\mathbf{x})} & \ge \mathbb{E}_{z\sim q_\phi(\mathbf{z}|\mathbf{x})}\left[ \log{p(\mathbf{z})} \right] + \mathbb{E}_{\mathbf{z}\sim q_\phi(\mathbf{z}|\mathbf{x})}\left[ \log{p_\theta(\mathbf{x}|\mathbf{z})} \right] - \mathbb{E}_{\mathbf{z}\sim q_\phi(\mathbf{z}|\mathbf{x})}\left[ \log{q_\phi(\mathbf{z}|\mathbf{x})} \right] \\
\label{eqn:vae6}
& = \mathbb{E}_{\mathbf{z}\sim q_\phi(\mathbf{z}|\mathbf{x})}\left[ \log{p_\theta(\mathbf{x}|\mathbf{z})} \right] - \mathbb{E}_{\mathbf{z}\sim q_\phi(\mathbf{z}|\mathbf{x})}\left[ \log{ \frac{q_\phi(\mathbf{z}|\mathbf{x})}{p(\mathbf{z})} } \right] \\
\label{eqn:vae7}
& = \mathbb{E}_{\mathbf{z}\sim q_\phi(\mathbf{z}|\mathbf{x})}\left[ \log{ p_\theta(\mathbf{x}|\mathbf{z}) } \right] - \mathcal{D}_{KL}( q_\phi(\mathbf{z}|\mathbf{x}) || p(\mathbf{z}) )
\end{align}

The right hand side of the equation \ref{eqn:vae7} is the ELBO of the observed distribution. It is compounded by two terms: the reconstruction likelihood of the data plus the Kullback-Leibler divergence ($\mathcal{D}_{KL}$) between two distributions; $\mathcal{D}_{KL}$ is also called relative entropy. Maximizing the ELBO ensures that we maximize the log-likelihood of the observed data.

From a practical point of view, VAEs learn to represent the images dataset on a low-dimensional manifold called latent space $Z$. The prior distribution of the latent space in most of the works, including here, is assumed to be a Multivariate Gaussian $N(0, 1)$ with diagonal covariance.

\subsection{Metrics}

When we reconstruct images, we need to use metrics that answer quantitatively how similar the generated and the reference images are, this approach is called Full Reference Image Quality Assessment (FR-IQA) \cite{wang2004image}. To accomplish FR-IQA, it is desirable metrics that match the Human Vision System (HVS), because the generated images should be appealing for the human eyes. The characteristics of the HVS for image quality perception are divided into four categories: contrast sensitivity function, luminance masking, contrast masking, and foveated masking \cite{seo2020novel}. 

Attempting to reach perceptual metrics, SSIM \cite{wang2004image} assumes that HVS is highly adapted for extracting structural information from the scene. This work was later extended to Multi-Scale Structural Similarity (MS-SSIM) \cite{wang2003multiscale} as SSIM depends on the right scale of the viewing conditions like display resolution and viewing distance; differently than MS-SSIM, which incorporates easier image details at different resolutions. MS-SSIM evaluates the image on an iterative process that applies a low-pass filter followed by a factor of two downsampling and a SSIM evaluation between the images; the process runs until the minimum desired scale is reached, the final result is a weighted summation of each scale result.

A newer approach on perceptual losses is the Learned Perceptual Image Patch Similarity (LPIPS) \cite{zhang2018unreasonable}, the authors argue that perceptual similarity is not a special function on its own, but rather a consequence of visual representations tuned to be predictive about important structures of the world. They extract deep features from a calibrated neural network and compare the distance between the resulting vectors of the image and its reference using an additional weighted trained layer. The training of the additional layer and posterior calibration of the pre-trained ones were made with their own dataset, Berkeley-Adobe Perceptual Patch Similarity (BAPPS), which counted with 484k images of human judgments labels.

Another paradigm, different than the perceptual approaches, was proposed by \cite{sheikh2006image}, the Visual Information Fidelity (VIF). VIF criterion is based on Natural Scene Statistics (NSS) and HVS. VIF quantifies the Shannon information present on the \emph{distorted} image, which is the reconstructed image, and the reference image itself. On the VIF comparison, each image is modeled as a Gaussian Scaled Mixture in the wavelet domain.

A more common choice in the computer vision community is $l_2$, usually the default choice. There are plenty of reasons to love $l_2$ as cited in \cite{wang2009mean}: it is simple, parameter-free, and cheap to compute. Besides, it satisfies convenient conditions of nonnegativity and identity symmetry; it has a clear physical meaning and often has a closed-form solution.

In the same $l_p$ norm family, we have $l_1$, which does not increase the punishment on larger errors much more than smaller errors as opposed to $l_2$, turning it more robust against outliers leverage. Moreover, as shown in the \cite{zhao2016loss} experiments, $l_1$ outperformed $l_2$ on the tasks of super-resolution, denoising, and JPEG artifacts removal; the authors hypothesized that $l_2$ gets stuck easier in a local minimum while $l_1$ can reach a better minimum.

\subsection{Related Works}

Previous works attempted to improve the image synthesis quality for VAE. 
\cite{hou2017deep} used a distinguish loss function, they employed a pre-trained VGG network to use the outcome of the 19th layer as a feature perceptual loss. \cite{qian2019make} explored a disentangled arrangement between concepts of appearance and structure, using a Conditional Variational Autoencoder (C-VAE), the structure was represented by a facial boundary map that comes from facial landmark interpolations; besides they adopted the idea of using Gaussian Mixture Models (GMM) on the latent space, turning it more complex than the usual single Gaussian distribution in the hope it would represent better face factors like ages, complexions, luminance, and poses.

Another common approach is the combination of VAEs and GANs concepts. For example, \cite{larsen2016autoencoding} appended a GAN discriminator at the VAE decoder, making it to learn an adversarial loss plus the reconstruction loss, the reconstruction loss was based on hidden feature maps of the middle layers in the discriminator net. The features from the GAN discriminator replaced the element-wise errors with feature-wise errors. \cite{khan2018adversarial} followed a similar approach, they appended a discriminator to the VAE and an adversarial loss to the objective function, their distinction was they choose an autoencoder as a discriminator.


Currently, the state-of-the-art solutions for generating high-fidelity images with VAE are VQ-VAE-2 \cite{razavi2019generating} and NVAE \cite{vahdat2020NVAE}. VQ-VAE-2 counted on a highly complex architecture that discretizes the latent space with Vector Quantization (VQ). The prototype vectors from the VQ codebook are learned throughout the training. Moreover, the architecture counted with hierarchical levels of representations and complex priors distributions of the latent space like self-attention layers.

NVAE used a deep hierarchical VAE with depth-wise separable convolutions and batch normalization. Besides, they proposed a residual parameterization approach to approximate the posterior parameters and to reduce the KL divergence; they also adopted spectral regularization to stabilize the training. They claimed that although VQ-VAE-2 used VAE, it did not optimize the ELBO of the data, which is the essential part of VAE's objective, thus this fact would make them the first successful VAE project applied to large images as 256x256 resolution.

\section{Proposed Solution}

\begin{figure}[ht]
\includegraphics[width=\textwidth]{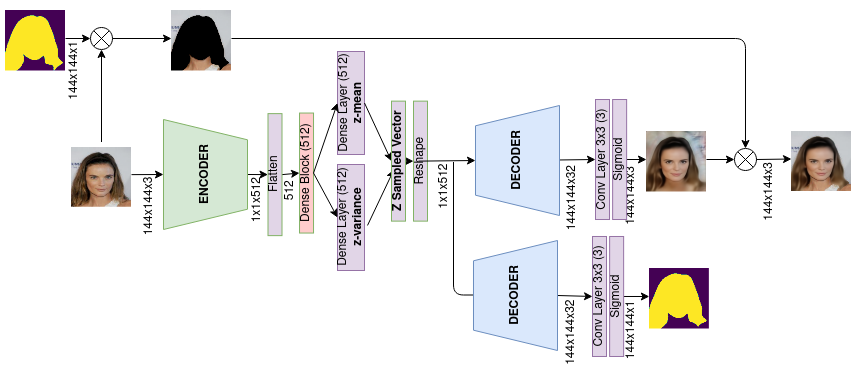}
\caption{Training mode architecture.} \label{fig:arch_training}
\end{figure}

Our proposed method is the addition of face masks into the architecture shown in Figure \ref{fig:arch_training}, which forces the losses functions of the neural network to be only impacted by the face pixels and to avoid any information from the background. During the training phase, the image background is replaced by the one of the original input.

The architecture includes an extra decoder responsible to predict a binary face mask for the corresponding predicted face in the first decoder. The training labels for this part comes from an external face segmentation model. During prediction, the background of the input image replaces the one from the predicted image by using the predicted mask. The mask decoder is trained with \emph{Binary Crossentropy} and \emph{Dice} losses.

\subsection{Architecture}

The encoder is built on multiple \emph{convolutional blocks}. A convolutional block consists of one convolutional layer of filter size 3x3 with "$n$" filters, batch normalization, and \emph{ELU} activation layer. The encoder employed a sequence of 2 blocks of 32 filters, 2 blocks of 64 filters, 2 blocks of 128 filters, 4 blocks of 256 filters, and 1 block of 512 filters. With 2D \emph{MaxPooling} layers of 2x2 among convolutional blocks to reduce the input dimensionality by half until the encoder reaches the dimension of 1x1x512.

The decoder had a symmetric reverse structure with the Encoder. \emph{Upsampling} was made by the convolutional transpose layers. The input shape used for this architecture was 144x144x3. In total, the NN summed 9.3M trained parameters. 

\subsection{Dataset}

The adopted dataset was CelebA \cite{liu2018large}. A largely referenced dataset on computer vision works related to face. It contains more than 200K images from 10K celebrity identities. Each image is annotated with 40 binary presence attributes like hair color, earrings, smiling, hat, and on.

\subsection{Experiments}

We realized an ablation study of ten hypotheses to measure the impact of the addition of the face mask and the usage of different losses functions. To accomplish it, we focused to answer 4 questions: Do face masks help? Which standalone loss does work better? Which $l_n$ does work better with SSIM? and finally, which of the ten hypotheses is the best?

Table \ref{tab:exp_hypothesis} summarizes each hypothesis. We tried the standalone losses functions SSIM, $l_1$, and $l_2$; afterwards, we combined $l_2$ and $l_1$ with SSIM. For each combination of losses functions, there is a version with and without face masks.

\begin{table}[h!]
\begin{adjustbox}{width=.5\columnwidth,center}
\centering
\begin{tabular}{|*{6}{P{1.5cm}|}}
    \hline
    & Mask & SSIM & $l_1$ & $l_2$ \\ [0.5ex] 
    \hline\hline
    $H_1$ & \checkmark & \checkmark & \checkmark & \\ \hline
    $H_2$ &  & \checkmark & \checkmark & \\ \hline
    $H_3$ & \checkmark &  & \checkmark & \\ \hline
    $H_4$ & & & \checkmark & \\ \hline
    $H_5$ & \checkmark & \checkmark & &  \\ \hline
    $H_6$ &  & \checkmark & &  \\ \hline
    $H_7$ & \checkmark & & & \checkmark  \\ \hline
    $H_8$ & & & & \checkmark  \\ \hline
    $H_9$ & \checkmark & \checkmark & & \checkmark  \\ \hline
    $H_{10}$ & & \checkmark & & \checkmark  \\ \hline
\end{tabular}
\end{adjustbox}
\caption{Hypotheses of the ablation study.}
\label{tab:exp_hypothesis}
\end{table}

All hypotheses used the same NN core architecture. All experiments ran 50 epochs of 4000 steps with a batch size of 32, using Adam optimizer with a learning rate of 1e-4 and gradient clipping normalization of 1e-3. The Kullback-leibler divergence loss was scaled by 1e-3 such it did not constrain too much the image reconstruction learning.

\section{Evaluation and Discussions}

The evaluation of each hypothesis was done by the metrics: SSIM, MS-SSIM, LPIPS, VIF, $l_1$, and $l_2$. SSIM, MS-SSIM, and VIF had their values inverted, subtracting the original quantities by 1, as  "1" is their maximum value. Then, for all metrics, the general rule is: the less the values are, the better the results are. $l_1$ and $l_2$ were measured in the 0-255 range, afterwards, they were scaled by the factor of `10 / 255`.

Only the face pixels were regarded for evaluation. This was accomplished by replacing the background of all predicted images, including the ones that the hypothesis model did not predict the face masks. The evaluation happened only on the test set images already defined on celebA dataset, which counts a total of 19962 images. Table \ref{tab:exp_results} summarizes the results:

\begin{table}[h!]
\begin{adjustbox}{width=.9\columnwidth,center}
\centering
\begin{tabular}{|*{7}{P{2.15cm}|}}
\hline
      & 1 - SSIM & 1 - MSSSIM & LPIPS & 1 - VIF & $l_1$ & $l_2$ \\ \hline\hline
$H_1$ & 0.311 & 0.082 & 0.145 & 0.529 & 0.508 & 0.732 \\ \hline
$H_2$ & 0.338 & 0.090 & 0.162 & 0.560 & 0.546 & 0.781 \\ \hline
$H_3$ & 0.343 & 0.095 & 0.151 & 0.541 & 0.534 & 0.774 \\ \hline
$H_4$ & 0.373 & 0.103 & 0.173 & 0.574 & 0.587 & 0.826 \\ \hline
$H_5$ & 0.321 & 0.093 & 0.146 & 0.537 & 0.585 & 0.836 \\ \hline
$H_6$ & 0.314 & 0.085 & 0.157 & 0.550 & 0.601 & 0.852 \\ \hline
$H_7$ & 0.322 & 0.082 & 0.146 & 0.527 & 0.501 & 0.694 \\ \hline
$H_8$ & 0.377 & 0.100 & 0.172 & 0.572 & 0.577 & 0.797 \\ \hline
$H_9$ & \textbf{0.307} & \textbf{0.077} & \textbf{0.143} & \textbf{0.523} & \textbf{0.487} & \textbf{0.682} \\ \hline
$H_{10}$ & 0.359 & 0.095 & 0.168 & 0.566 & 0.562 & 0.782 \\ \hline
\end{tabular}
\end{adjustbox}
\caption{Hypotheses results. The best performance for each metric is in bold.}
\label{tab:exp_results}
\end{table}

Now we intend to answer the four questions raised in the Experiments subsection.

\subsection{Do face masks help?}

\begin{figure}[ht]
\includegraphics[width=\textwidth]{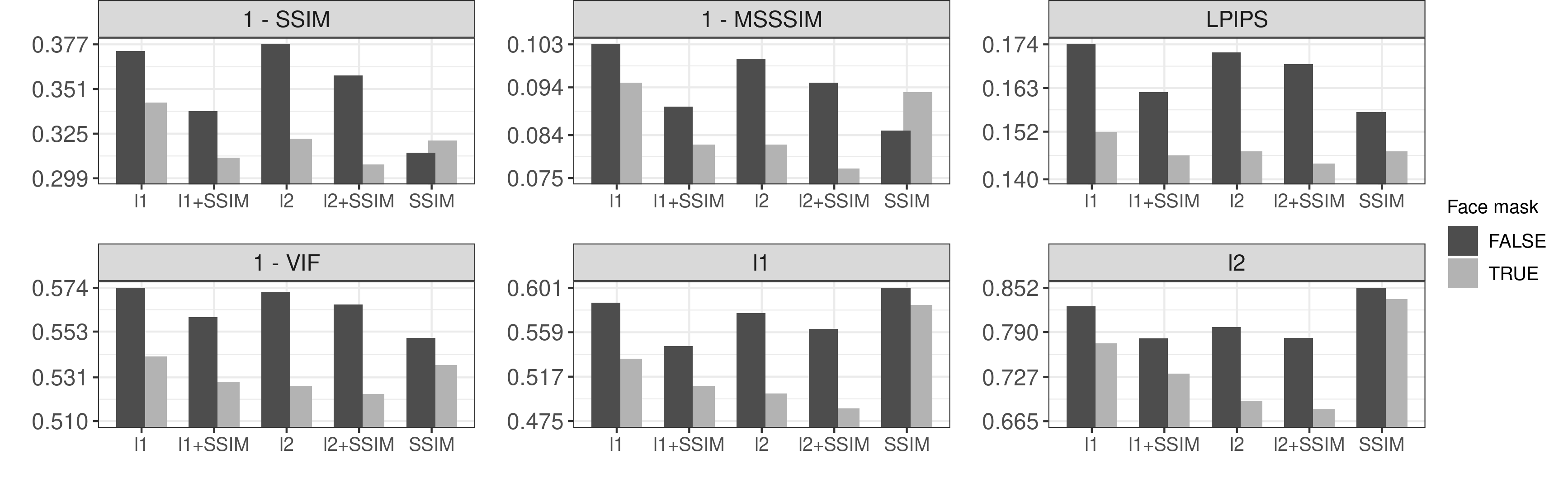}
\caption{Evaluation of the face mask impact. On the top row are the perceptual metrics.} \label{fig:results_maskbased}
\end{figure}

Yes, it was well noticed in Figure \ref{fig:results_maskbased} that the addition of the face masks enhanced all metrics practically for all losses combination, but the SSIM standalone hypotheses (H5/H6), where the without mask case performed better on the SSIM and MS-SSIM metrics. It's counterintuitive from what we expected, however, if we regard LPIPS as a better perceptual assessment than the SSIM ones, the hypothesis with face mask kept working better on H5/H6 pair case.

While the addition of face masks was very effective for the $l_n$ losses, the SSIM standalone cases proved less sensitive for the background information, and the application of the masks did not change the results on their cases.

\subsection{Which standalone loss does work better?}

\begin{figure}[ht]
\includegraphics[width=\textwidth]{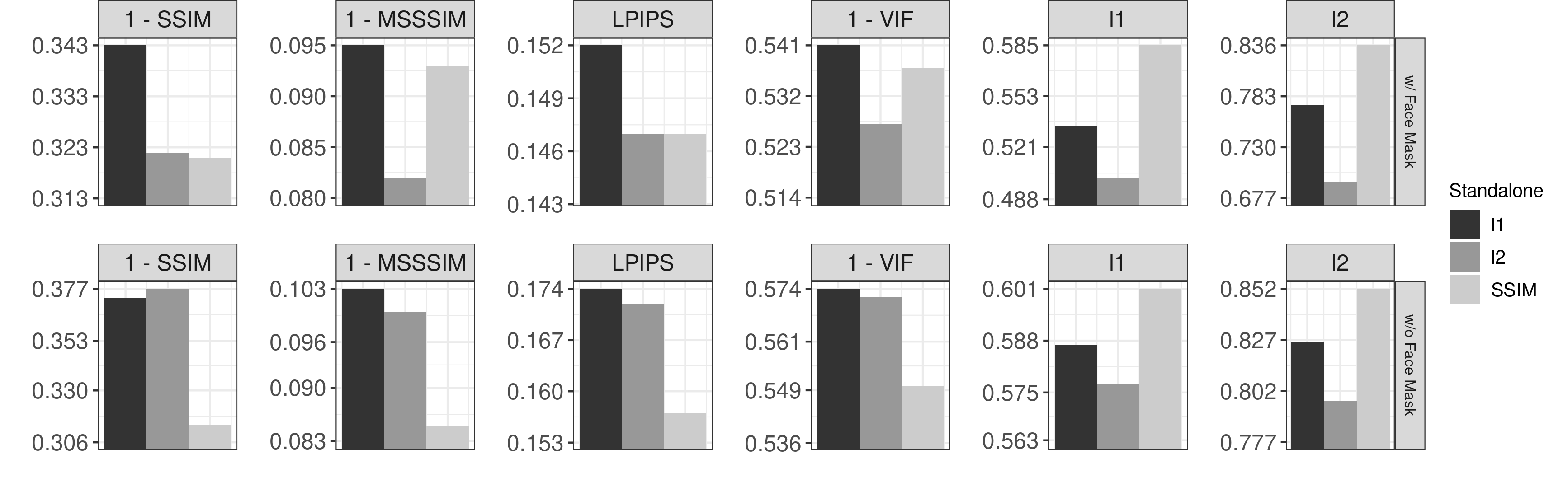}
\caption{Evaluation of the effect of each loss applied alone. Face masks hypothesis results are on the top row, and the without ones are on the bottom row.} \label{fig:results_standaloneloss}
\end{figure}

In Figure \ref{fig:results_standaloneloss}, looking at the scenario with face masks, we were surprised how well $l_2$ did on perceptual metrics (SSIM, MS-SSIM, and LPIPS), being better or equal to SSIM cases; besides $l_2$ outperformed SSIM on the other metrics. On the other hand, when not using face masks, $l_2$ performed way worse than SSIM for the perceptual assessments.  

An interesting fact is that $l_2$ outperformed $l_1$ even on the $l_1$ metric. The takeaways found here are: $l_2$ usually performs better than $l_1$ on the face reconstruction task; the usage of face masks improves a lot the performance of $l_2$, working better or equal the SSIM standalone. But, without the mask addition, SSIM is way better on the perceptual evaluation.

\subsection{Which $l_n$ does work better with SSIM?}

It is known that SSIM was made to work with grayscale images, and it may present problems with color images \cite{zhao2016loss, nilsson2020understanding}. To improve the training process with the perceptual loss SSIM, we added a $l_n$ loss to complement SSIM.

\begin{figure}[ht]
\includegraphics[width=\textwidth]{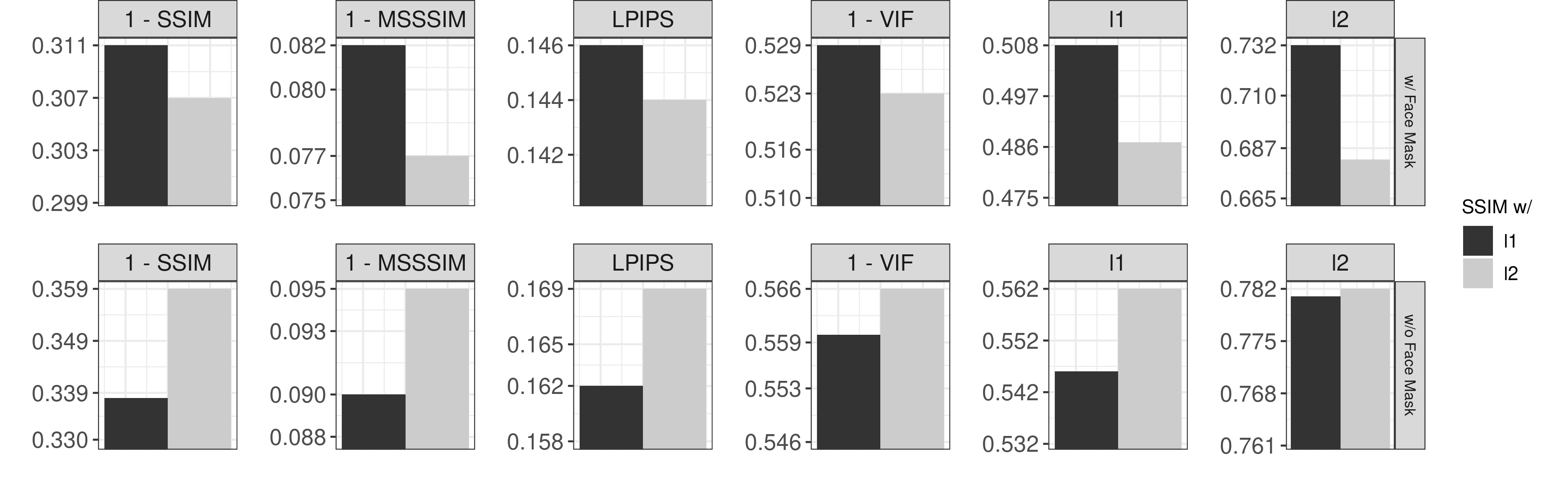}
\caption{Evaluation of the effect of each loss applied alone. Hypotheses with face masks are on the top row, and the without ones are on the bottom row.} \label{fig:results_l1orl2}
\end{figure}

Figure \ref{fig:results_l1orl2} showed up two very distinct scenarios. Using the mask approach, $l_2$ clearly surpasses $l_1$ in all metrics, but without the mask, $l_1$ obtained the best results.

Probably, here is a case where the higher sensitivity of $l_2$ to outliers played badly. We assume the amount of information on the background yielded a higher rate of errors on $l_2$, attenuating the right gradients for the main task of face reconstruction.

\subsection{Which hypothesis is the best?}

We had the dominant hypothesis $H_9$ pointed in Table \ref{tab:exp_results}, which reached the best performance for all metrics. $H_9$ was compounded by face mask, SSIM, and $l_2$ loss.

Among the "without masks" hypotheses, it did not show up any dominant hypothesis. On the perceptual metrics, $H_5$ worked better with the SSIM standalone, and with regards to the $l_n$ metrics, $H_1$, which is SSIM + $l_1$ was the best.

\subsection{Visual comparisons}

\begin{figure}[ht]
\includegraphics[width=\textwidth]{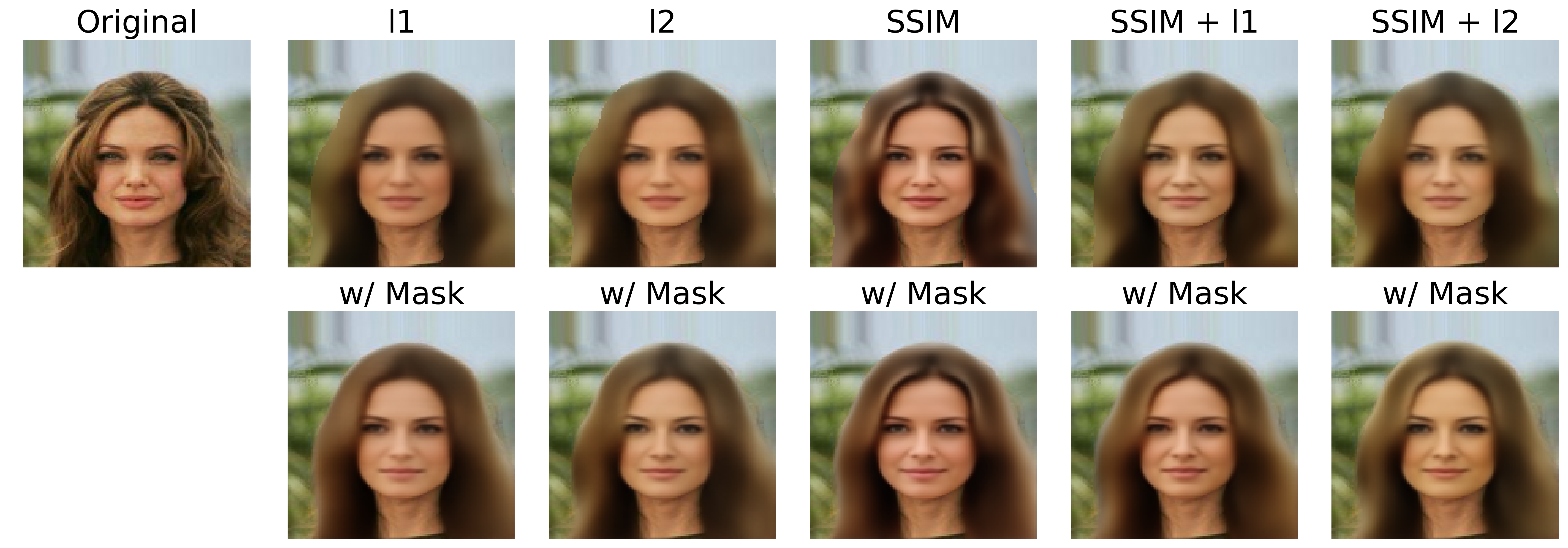}
\caption{Visual comparison between hypotheses.} \label{fig:visual_1}
\end{figure}

All images presented here are from the celebA test set. Checking out the hypotheses outcomes, we noticed some patterns. Looking at Figure \ref{fig:visual_1}, we see the blurred reconstructed hair, which is a recurrent issue with VAE. Although, it is possible to notice that the hypotheses with face masks got slightly sharper lines on the hair, especially SSIM standalone.

The SSIM standalone clearly had the crispest reconstructed samples, however, the colors presented on the face are the most distant from the reference image, even they appearing brighter, they moved away from the original yellowish color. The hypothesis of SSIM + $l_2$ with face masks offered a more fidelity outcome.

\begin{figure}[ht]
\includegraphics[width=\textwidth]{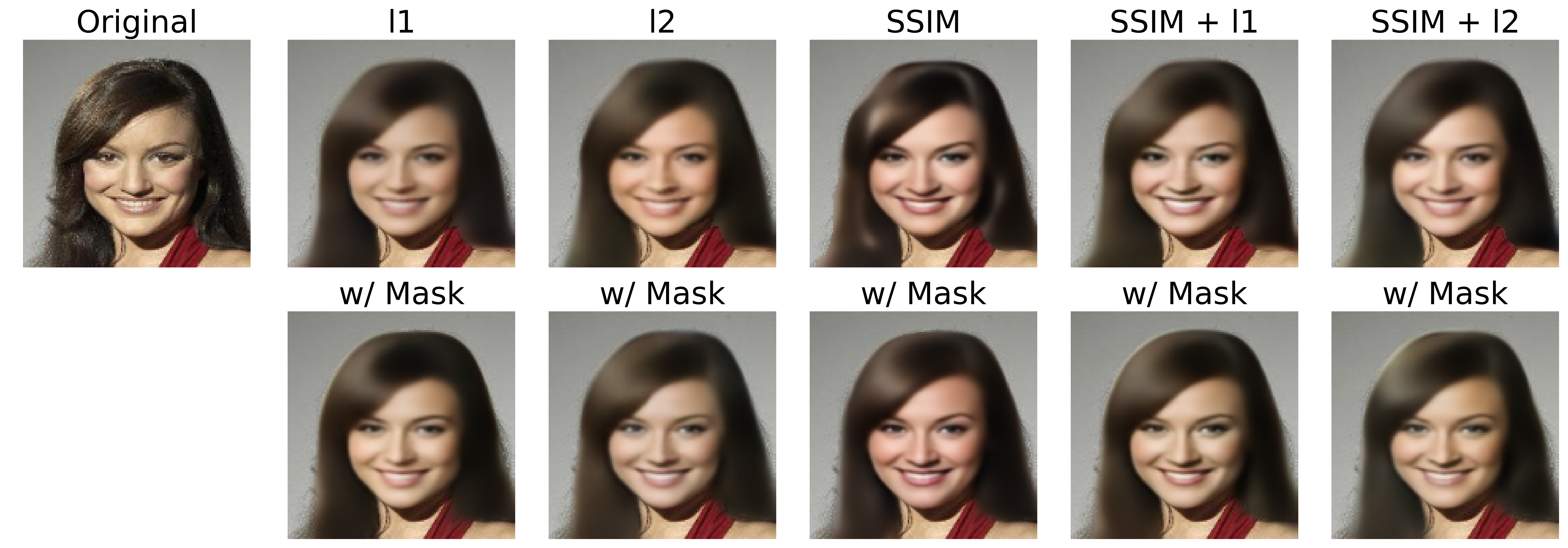}
\caption{Visual comparison between hypotheses.} \label{fig:visual_2}
\end{figure}

In Figure \ref{fig:visual_2}, we noticed crisper samples in the SSIM hypotheses, but its color again is shifted, moving away from the greenish presence of the original picture. The hypotheses with face masks preserved more the greenish, and also the teeth presence is better replicated. In this figure, the hypothesis with SSIM + $l_1$ + face masks presented the best reconstruction of the original image.

\begin{figure}[ht]
\includegraphics[width=\textwidth]{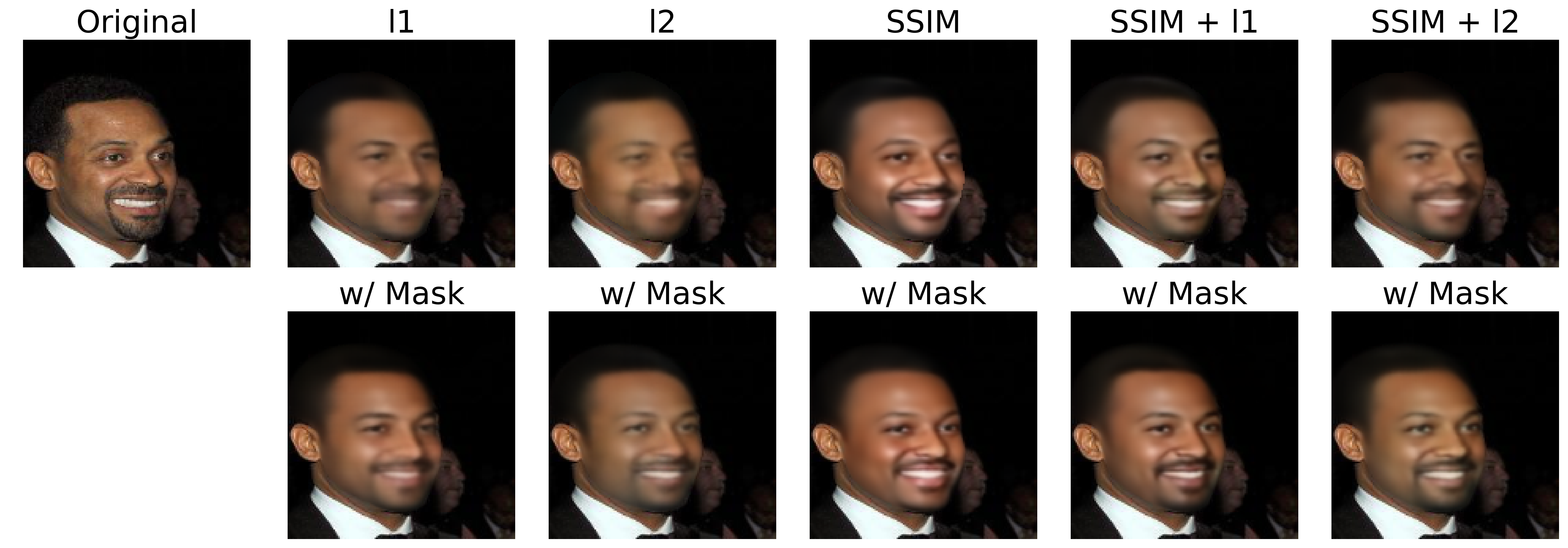}
\caption{Visual comparison between hypotheses.} \label{fig:visual_4}
\end{figure}

Figure \ref{fig:visual_4} had a posing challenge. Clearly, the usage of the face mask or SSIM was relevant to accomplish a good face reconstruction. Again only SSIM changed the original color image, making the skin quite reddish; SSIM + $l_1$ showed a more proper solution.

\section{Conclusions}

We saw that a simple alternative like the addition of a face mask branch on VAE enhances the face reconstruction performance. Although it is not enough to yield crisp images as the current GANs' state-of-the-art works, it adds one more possibility towards better VAEs' frameworks.

From the experiments, we noted that SSIM is the loss function between the popular choices that yield the crispest samples, however, the rendered colors are usually shifted from the original image. The addition of $l_n$ to SSIM helps to regulate the color shifting, in this case, when the face masks are used $l_2$ is the best option, otherwise, $l_1$ is the best. 

The face masks were especially effective when any $l_n$ was used. Observing only the standalone cases, $l_2$ was equally competitive to SSIM on perceptual assessments when face masks were used.

SSIM standalone was the only case that face masks did not have an evident effect, the background information does not seem to disturb the face reconstruction learning when SSIM standalone loss is applied. On the opposite, $l_n$ options showed very sensitivity to the presence of background information.

\section*{Future Work}

The analysis presented here can be extended to the edition and manipulation of facial attributes, to the synthesis of random samples created from random sampling on the latent space, and to the comparison against state-of-the-art works from VAEs and GANs using Fréchet Inception Distance (FID) and Inception Score (IS).

\bibliographystyle{sbc}
\bibliography{main}

\end{document}